\documentclass[letterpaper, conference]{IEEEtran}

\IEEEoverridecommandlockouts                              

\usepackage{subcaption}
\usepackage{hyperref}       
\usepackage{booktabs}       
\usepackage{graphicx}
\usepackage[acronym,shortcuts]{glossaries}
\usepackage{comment}
\usepackage{enumerate}
\usepackage[capitalise]{cleveref}
\usepackage[table,xcdraw]{xcolor}
\usepackage{siunitx}
\usepackage{bm} 
\usepackage{amssymb} 
\usepackage{amsmath}
\usepackage[english]{babel}
\usepackage{adjustbox}
\usepackage{multirow}
\usepackage{pifont}
\usepackage{cite}

\definecolor{lightgray}{gray}{0.9}

\def\BibTeX{{\rm B\kern-.05em{\sc i\kern-.025em b}\kern-.08em
    T\kern-.1667em\lower.7ex\hbox{E}\kern-.125emX}}
\usepackage{eso-pic}

\newcommand\AtPageUpperCenterNotice[1]{%
  \AtPageUpperLeft{%
    \put(\LenToUnit{0.5\paperwidth},\LenToUnit{-2cm}){\makebox[0pt]{#1}}%
  }%
}

\AddToShipoutPictureBG*{%
  \AtPageUpperCenterNotice{%
    \parbox[b][0.5cm][c]{\paperwidth}{%
      \centering
      \fontsize{12}{14}\selectfont
      \color{gray!50}
      This paper has been accepted for publication in\\
       International Conference on Intelligent Robots and Systems (IROS), Hangzhou 2025. \copyright{}IEEE.
    }%
  }%
}

\AddToShipoutPictureBG*{
  \AtPageLowerLeft{%
    \raisebox{25pt}{\makebox[\paperwidth]{\begin{minipage}{21cm}\centering
    \fontsize{10}{8}\selectfont
    \textcolor{gray!50}{%
      \copyright{} 2025 IEEE. Personal use of this material is permitted.
      Permission from IEEE must be obtained for all other uses, in any current or future media,
      including reprinting/republishing this material for advertising or promotional purposes,
      creating new collective works, for resale or redistribution to servers or lists,
      or reuse of any copyrighted component of this work in other works.
    }
    \end{minipage}}}%
  }
}

\title{\LARGE \bf \vspace{6mm}
$\mathcal{M}$-Predictive Spliner: Enabling Spatiotemporal Multi-Opponent Overtaking for Autonomous Racing 
}

\author{Nadine Imholz\IEEEauthorrefmark{1}, Maurice Brunner\IEEEauthorrefmark{1}, Nicolas Baumann\IEEEauthorrefmark{1}, Edoardo Ghignone\IEEEauthorrefmark{1}, and Michele Magno\IEEEauthorrefmark{1} \\
\thanks{\IEEEauthorrefmark{1}Nadine Imholz, Maurice Brunner, Nicolas Baumann, Edoardo Ghignone, and Michele Magno are associated with the Center for Project-Based Learning, D-ITET, ETH Zurich.}%
\thanks{Corresponding author: Nicolas Baumann.}
}

\begin{document}
\newacronym{lidar}{LiDAR}{Light Detection and Ranging}
\newacronym{radar}{RADAR}{Radio Detection and Ranging}
\newacronym{ml}{ML}{Machine Learning}
\newacronym{if}{IF}{Intermediate Frequency}
\newacronym{aoa}{AoA}{Angle of Arrival}
\newacronym{cfar}{CFAR}{Constant False Alarm Rate}
\newacronym{iot}{IoT}{Internet of Things}
\newacronym{bev}{BEV}{Bird's-Eye View}
\newacronym{sota}{SotA}{State-of-the-Art}
\newacronym{cr3dt}{CR3DT}{Camera-RADAR 3D Detection and Tracking}
\newacronym{cr3d}{CR3D}{Camera-RADAR 3D Detector}
\newacronym{cnn}{CNN}{Convolutional Neural Network}
\newacronym{lss}{LSS}{Lift Splat Shoot}
\newacronym{rgb}{RGB}{Red Green Blue}
\newacronym{map}{mAP}{mean Average Precision}
\newacronym{fpn}{FPN}{Feature Pyramid Network}
\newacronym{iou}{IOU}{Intersection over Union}
\newacronym{nds}{NDS}{nuScenes Detection Score}
\newacronym{mate}{mATE}{mean Average Translation Error}
\newacronym{fov}{FoV}{Field of View}
\newacronym{vfe}{VFE}{Voxel Feature Encoding}
\newacronym{amota}{AMOTA}{Average Multi-Object Tracking Accuracy}
\newacronym{amotp}{AMOTP}{Average Multi-Object Tracking Precision}
\newacronym{mot}{MOT}{Multi-Object Tracking}
\newacronym{ids}{IDS}{ID Switches}
\newacronym{roi}{RoI}{Region of Interest}
\newacronym{nms}{NMS}{Non-Maximum Suppression}
\newacronym{mave}{mAVE}{mean Average Velocity Error}
\newacronym{cbgs}{CBGS}{Class-Balanced Grouping and Sampling}
\newacronym{gpu}{GPU}{Graphics Processing Unit}
\newacronym{fps}{FPS}{Frames Per Second}
\newacronym{kf}{KF}{Kalman Filter}
\newacronym{pspliner}{PSpliner}{Predictive Spliner}
\newacronym{mpspliner}{$\mathcal{M}$-PSpliner}{Multi Opponent Predictive Spliner}
\newacronym{ftg}{FTG}{Follow The Gap}
\newacronym{gbo}{GBO}{Graph-Based Overtake}
\newacronym{gp}{GP}{Gaussian Process}
\newacronym{rbf}{RBF}{Radial Basis Function}
\newacronym{sqp}{SQP}{Sequential Quadratic Programming}
\newacronym{roc}{RoC}{Region of Collision}
\newacronym{cpu}{CPU}{Central Processing Unit}
\newacronym{cots}{CotS}{Commercial off-the-Shelf}
\newacronym{obc}{OBC}{On-Board Computer}
\newacronym{mpc}{MPC}{Model Predictive Control}
\newacronym{tpr}{TPR}{True Positive Rate}
\newacronym{fdr}{FDR}{False Detection Rate}
\newacronym{rmse}{RMSE}{Root Mean Squared Error}
\newacronym{sort}{SORT}{Simple, Online, and Realtime Tracking}
\newacronym{cv}{CV}{Computer Vision}
\newacronym{id}{ID}{IDentity}
\newacronym{reid}{reID}{Re-Identification}
\newacronym{abd}{ABD}{Adaptive Breakpoint Detector}
\newacronym{gpr}{GPR}{Gaussian Process Regression}
\newacronym{los}{LoS}{Line of Sight}

\maketitle
\thispagestyle{empty}
\pagestyle{empty}

\begin{abstract}  
Unrestricted multi-agent racing presents a significant research challenge, requiring decision-making at the limits of a robot's operational capabilities. While previous approaches have either ignored spatiotemporal information in the decision-making process or been restricted to single-opponent scenarios, this work enables arbitrary multi-opponent head-to-head racing while considering the opponents' future intent. The proposed method employs a \gls{kf}-based multi-opponent tracker to effectively perform opponent \gls{reid} by associating them across observations. Simultaneously, spatial and velocity \gls{gpr} is performed on all observed opponent trajectories, providing predictive information to compute the overtaking maneuvers. This approach has been experimentally validated on a physical 1:10 scale autonomous racing car achieving an overtaking success rate of up to 91.65\% and demonstrating an average 10.13\%-point improvement in safety at the same speed as the previous \gls{sota}. These results highlight its potential for high-performance autonomous racing.  
\end{abstract}  
\glsresetall

\setlength{\tabcolsep}{3pt}
\section{Introduction}
Autonomous racing serves as a crucial proving ground for broader autonomous driving technologies, facilitating the rigorous testing of algorithms and hardware under extreme operational conditions \cite{betz_weneed_ar, ar_survey, forzaeth}. Within this field, unrestricted multi-opponent head-to-head racing represents a research challenge, necessitating advanced multi-agent decision-making and motion planning at high velocities.

Local planners play a primary role in such races \cite{forzaeth, pspliner}, and their role is crucial to enabling overtaking maneuvers on a constrained track. The development of these overtaking strategies focuses on creating robust methods that simplify operations and minimize the need for frequent heuristic adjustments concerning the opponent's behavior. However, this simplicity often conflicts with the parameter sensitivity and the computational depth required by model-based techniques \cite{mpcc}, alongside the intricate heuristics essential for reactive \cite{ftg, forzaeth}, sampling \cite{frenetplanner}, and graph-based \cite{gbo} overtaking algorithms. While these methods predominantly account for the spatial positioning of opponents, few algorithms integrate this data in a spatiotemporal context, such as \gls{pspliner} \cite{pspliner}. A limitation of \gls{pspliner}, however, is its restriction to scenarios involving only a single opponent, thereby inadequately addressing genuine multi-opponent racing scenarios.

To overcome these limitations, \gls{mpspliner} proposes an advancement over the single-opponent-focused \gls{pspliner}. It enhances the multi-opponent tracking and perception system to accommodate an arbitrary number of competitors, utilizing \emph{Norfair} \cite{norfair} and custom adaptations for racing contexts. This system regresses the trajectories of multiple opponents similar to \gls{pspliner}, using this comprehensive trajectory data to formulate spatiotemporal multi-opponent overtaking strategies. These strategies have been demonstrated to operate on computationally limited robotic processing platforms and validated using 1:10 scale autonomous racing vehicles. To summarize the contributions of this work:
\begin{enumerate}[I]
    \item \textbf{Multi-Opponent Capability:} This work enables fully unrestricted multi-opponent capability in head-to-head racing, leveraging spatiotemporal information of opponents. It does so by leveraging a \gls{kf} that can track multiple vehicles via effective \gls{reid}, and by estimating and predicting the opponents' trajectories via \gls{gpr}. The \gls{mpspliner} not only handles multiple opponents effectively but also matches the \gls{pspliner} \gls{sota} baseline in scenarios involving a single opponent. \gls{mpspliner} achieves successful overtaking at up to 76.75\% of the \textit{ego} vehicle's speed and demonstrates a 91.65\% success rate against single opponents. When tested with two opponents, the cumulative time required for overtaking remains nearly identical to the single-opponent case, confirming its capabilities in a multi-opponent case.
    \item \textbf{In-Field Validation:} The \gls{mpspliner} framework is empirically validated using a 1:10 scale autonomous racing vehicle equipped with commercial off-the-shelf components and open-source software \cite{forzaeth}. The vehicle's main processing unit is an Intel NUC \texttt{i7-1165G7}, and it utilizes a Hokuyo UST-LX10 \gls{lidar} for environmental sensing, ensuring reproducibility of the results.
    \item \textbf{Open-Source Commitment:} This research is based on entirely open-source software \cite{forzaeth, pspliner} and hardware \cite{okelly2020f1tenth, forzaeth}. Furthermore, everything in this project is fully open-sourced, allowing the research community to build upon and extend our work. The code is available at: \href{github.com/ForzaETH/notavailableyet}{\url{github.com/ForzaETH/notavailableyet}}.
\end{enumerate}

\section{Related Work}
This section provides an overview of object tracking methods in \gls{cv} and general tracking approaches (\Cref{subsec:rw_tracking}). Additionally, commonly used overtaking planners in autonomous racing are reviewed (\Cref{subsec:rw_ot}).

\subsection{Tracking}\label{subsec:rw_tracking}
Tracking extends detection by maintaining the temporal identity of detected objects. In \gls{cv}, various tracking methodologies have been explored. Early approaches, predating \gls{ml}, relied on image-based feature matching techniques, such as Boosting, KCF, and MedianFlow \cite{boosting, kcf, medianflow}. These trackers operate purely on image data, leveraging classical \gls{cv} techniques, and depend on an object detector to first identify the \gls{roi}. However, trackers solely relying on image features, tend to be unreliable, often losing track of the target over time, particularly under occlusions \cite{xiao2024motiontrack}.

An alternative approach is to use a \gls{kf} to track the centroid of the detected \gls{roi} over time, as implemented in the \gls{sort} algorithm \cite{sort}. This method maintains object identities even during occlusions by utilizing a motion model, where tracklet matching and data association are performed via the \emph{Hungarian} algorithm, enabling \gls{mot}. The \textit{Deep}-\gls{sort} extension further enhances this by incorporating image embeddings, improving \gls{reid} over longer occlusions \cite{sort}.

In this work, the \emph{Norfair} tracker is utilized, a software package implementing functionalities similar to \gls{sort} and \textit{Deep}-\gls{sort}, while allowing full customization of the data association process and \gls{id} assignment \cite{norfair}. While \emph{Norfair} is typically applied in \gls{cv} for image-based tracking, we modify its implementation following \cite{davide} to operate outside the image domain. Since \emph{Norfair} is fundamentally built around a \gls{kf}, it can be adapted to spatial tracking by using real-world position data instead of pixel coordinates, enabling effective tracking in a spatial domain.

\subsection{Overtaking Algorithms}\label{subsec:rw_ot}
In scaled autonomous racing, the \gls{ftg} overtaking planner \cite{ftg} is a widely used reactive approach that does not rely on global positioning but instead processes \gls{lidar} data to follow the largest observed \gls{lidar} path. However, this method is suboptimal as it cannot adhere to a globally optimized racing trajectory and depends heavily on heuristics and manual tuning.

An alternative is the \emph{Spliner} method \cite{forzaeth}, which reacts to an opponent by generating a spline around it that rejoins the racing line. While simple, this approach is limited to a single opponent and still incorporates heuristic-based assumptions about opponent behavior.

Sampling-based methods, such as \emph{Frenet} planners \cite{frenetplanner}, generate multiple candidate trajectories in parallel, each assigned a cost function. The trajectory with the lowest cost is selected. However, defining appropriate cost metrics remains challenging and is often heuristic-driven, particularly when characterizing good overtaking behavior.

Graph-based methods, such as \gls{gbo} \cite{gbo}, frame the overtaking problem as a graph-search problem. A static lattice is embedded into the track, with transitions between nodes assigned costs. Standard graph search algorithms, such as \emph{Dijkstra's} algorithm, can then determine feasible overtaking paths, even with multiple opponents. However, this approach also relies heavily on heuristics, as vehicle dynamics and opponent behavior must be manually incorporated into the graph representation.

\gls{mpc}-based overtaking solutions integrate true vehicle dynamics by enforcing opponent-aware boundary constraints \cite{mpcot}. These methods theoretically enable multi-opponent overtaking but suffer from sensitivity to model accuracy. Notably, tire dynamics—critical in autonomous racing --- are notoriously difficult to estimate and model correctly \cite{story_of_modelmismatch}. Due to this complexity, \gls{mpc} is rarely used in scaled autonomous racing \cite{forzaeth}. Additionally, opponent behavior, which shapes the boundary constraints, remains heuristic-driven unless predictive opponent trajectory estimation is incorporated \cite{pspliner}.

\gls{pspliner} improves upon prior methods by explicitly predicting opponent behavior. It leverages a \gls{gp} to model the opponent’s trajectory, enabling the planner to compute an overtaking trajectory that accounts for a future \gls{roc}. This eliminates the need for heuristic-based opponent modeling but is limited to a single opponent.

In this work, we extend \gls{pspliner} to handle multi-opponent scenarios, introducing \gls{mpspliner}. This method integrates a multi-opponent tracker and extends \gls{pspliner} to consider the \glspl{roc} of multiple opponents. By incorporating opponent behavior predictions into a multi-agent framework, \gls{mpspliner} balances computational efficiency with robust multi-opponent overtaking capabilities, as summarized in \Cref{tab:overtaking_algorithms}.

\begin{table}[!htb] 
    \centering 
    \resizebox{\columnwidth}{!}{%
    \begin{tabular}{l|c|c|c|c}
    \toprule
    \textbf{Algorithm} & \textbf{Multi. Opp.} & \textbf{Temporal} & \textbf{Opp. Heuristics} &  \textbf{Compute [\%]$\downarrow$} \\
    \midrule
    \acrshort{ftg} \cite{ftg} & \textbf{Yes} & No & High & 17.6 \\
    Frenet \cite{frenetplanner} & \textbf{Yes} & No & High & 40.3 \\
    Spliner \cite{forzaeth} & No & No & High & \textbf{6.1} \\
    \acrshort{gbo} \cite{gbo} & \textbf{Yes} & No & High & 56.4 \\
    \acrshort{pspliner} \cite{pspliner} & No & \textbf{Yes} & \textbf{Low} & 23.9  \\
    \acrshort{mpspliner} \textbf{(ours)} & \textbf{Yes} & \textbf{Yes} & \textbf{Low} & 31.0  \\
    \bottomrule
    \end{tabular}%
    }
    \caption{Comparison of overtaking planners based on multi-opponent handling (desired: yes), temporal awareness of the opponent (desired: yes), reliance on ego/opponent heuristics (desired: low), and computational cost (desired: low) is the \gls{cpu} utilization rate from \Cref{tab:compute}.}
    \label{tab:overtaking_algorithms}
\end{table}

\section{Methodology}\label{sec:method}
This section details the implementation of the multi-opponent tracker with its \gls{reid} capabilities in \Cref{subsec:meth_tracker}. Additionally, \Cref{subsec:meth_prediction} outlines the mechanisms required to integrate multiple opponents into the overtaking strategy, ensuring spatiotemporal consideration for effective planning. Both the baseline \gls{pspliner} and the proposed \gls{mpspliner} are implemented into the 1:10 scaled open-source autonomous racing stack \emph{ForzaETH} to facilitate reproducibility \cite{forzaeth}.

\subsection{Multi-Opponent Tracker}\label{subsec:meth_tracker}
The detection measurements used to update the \gls{kf} come from a \gls{lidar}-based \gls{abd}  \cite{amin2022, forzaeth}, which segments raw \gls{lidar} scans into obstacle clusters. This approach identifies discontinuities in consecutive \gls{lidar} points to separate objects from the track. Detection and tracking are conducted in the robot's local Cartesian coordinate frame, where the \( x \)-axis points forward and the \( y \)-axis extends to the left.

To reduce false positives, clusters overlapping with the known track boundaries are removed using a curvilinear \textit{Frenet} transformation. Additionally, objects below a size threshold are discarded to further enhance detection accuracy. The remaining clusters are fitted with rectangles in 2D space, providing position, orientation, and size information essential for tracking and planning. The final detection achieves a high \gls{tpr} of 97\% while maintaining a low \gls{fdr} of 2\%, with a positional detection error of \( \bm{\mu}_{\text{err}} = (-0.08 \text{ m}, 0.01 \text{ m}) \) in the \( x \) and \( y \) \cite{forzaeth}.

These detections are then used within the image-based \emph{Norfair} tracker, adapted for spatial tracking as described in \cite{davide}, which employs a \gls{kf} with a point-mass model under a constant velocity assumption. The \gls{kf} system is defined as:

\begin{equation}
    \mathbf{x} = \begin{bmatrix} \mathbf{p} \\ \mathbf{v} \end{bmatrix}, \quad
    \mathbf{p} = \begin{bmatrix} x \\ y \end{bmatrix}, \quad
    \mathbf{v} = \begin{bmatrix} v_x \\ v_y \end{bmatrix}, \quad
    \mathbf{I} \in \mathbb{R}^{2 \times 2}
\end{equation}

\begin{equation}
    \mathbf{F} = \begin{bmatrix} 
        \mathbf{I} & \Delta t \mathbf{I} \\ 
        \mathbf{0} & \mathbf{I} 
    \end{bmatrix}, \quad
    \mathbf{H} = \begin{bmatrix} \mathbf{I} & \mathbf{0} \end{bmatrix} \in \mathbb{R}^{2 \times 4}, \quad
    \mathbf{R} = r \mathbf{I}
\end{equation}

\begin{equation}
    \mathbf{P}_0 = 
    \begin{bmatrix}
        \sigma^2_{\text{pos}} \mathbf{I} & \sigma_{\text{pos-vel}} \mathbf{I} \\
        \sigma_{\text{pos-vel}} \mathbf{I} & \sigma^2_{\text{vel}} \mathbf{I}
    \end{bmatrix}, \quad
    \mathbf{Q} = q \mathbf{I}, \quad
    \mathbf{z} = \begin{bmatrix} z_x \\ z_y \end{bmatrix} \in \mathbb{R}^{2}
\end{equation}

where \( \mathbf{p} \) and \( \mathbf{v} \) represent position and velocity, respectively, and \( \mathbf{z} \) denotes the positional detections provided by the perception system of \cite{forzaeth}. The initial covariance \( \mathbf{P}_0 \) and noise parameters are set as \( \sigma^2_{\text{pos}} = 10 \), \( \sigma_{\text{pos-vel}} = 0 \), \( \sigma^2_{\text{vel}} = 1 \), \( q = 1.7 \times 10^{-4} \), and \( r = 0.074 \).

The tracking process associates detections with existing \gls{kf}-tracklets \( \mathbf{T_j} \in \mathbb{R}^{4}\) from the aforementioned \gls{kf} tracker, using a \gls{reid} approach within a \gls{sort}-style framework, minimizing the Euclidean distance for assignment. The association process consists of the following steps:

\begin{equation}
    \begin{aligned}
        &\; \mathcal{M} = \arg \min_A \sum_{(i,j) \in A} \| \mathbf{z_i} - \mathbf{H T_j} \|_2, \\
        &\; \mathbf{T_j} \gets \texttt{KFUpdate}(\mathbf{T_j}, \mathbf{z_i}) \quad \forall (i,j) \in \mathcal{M}.
    \end{aligned}
\end{equation}

Matched tracklets are updated using the \gls{kf}, while unmatched tracklets undergo \gls{reid} based on the distance function \( d_{\text{ReID}}(\mathbf{T_j}, \mathbf{z_k}) \) as defined in \Cref{eq:reid}. The assignment set \( A \) consists of all feasible detection to tracklet associations that minimize the total distance in the association step. A tracklet is re-associated and a tracklet counter is incremented with a detection if the re-identification distance satisfies:

\begin{equation}\label{eq:reid}
    d_{\text{ReID}}(\mathbf{T_j}, \mathbf{z_k}) < \tau_{\text{ReID}}, \quad \tau_{\text{ReID}} = 0.1.
\end{equation}

\begin{figure}[!htb]
    \centering
    \includegraphics[angle=0, width=\columnwidth]{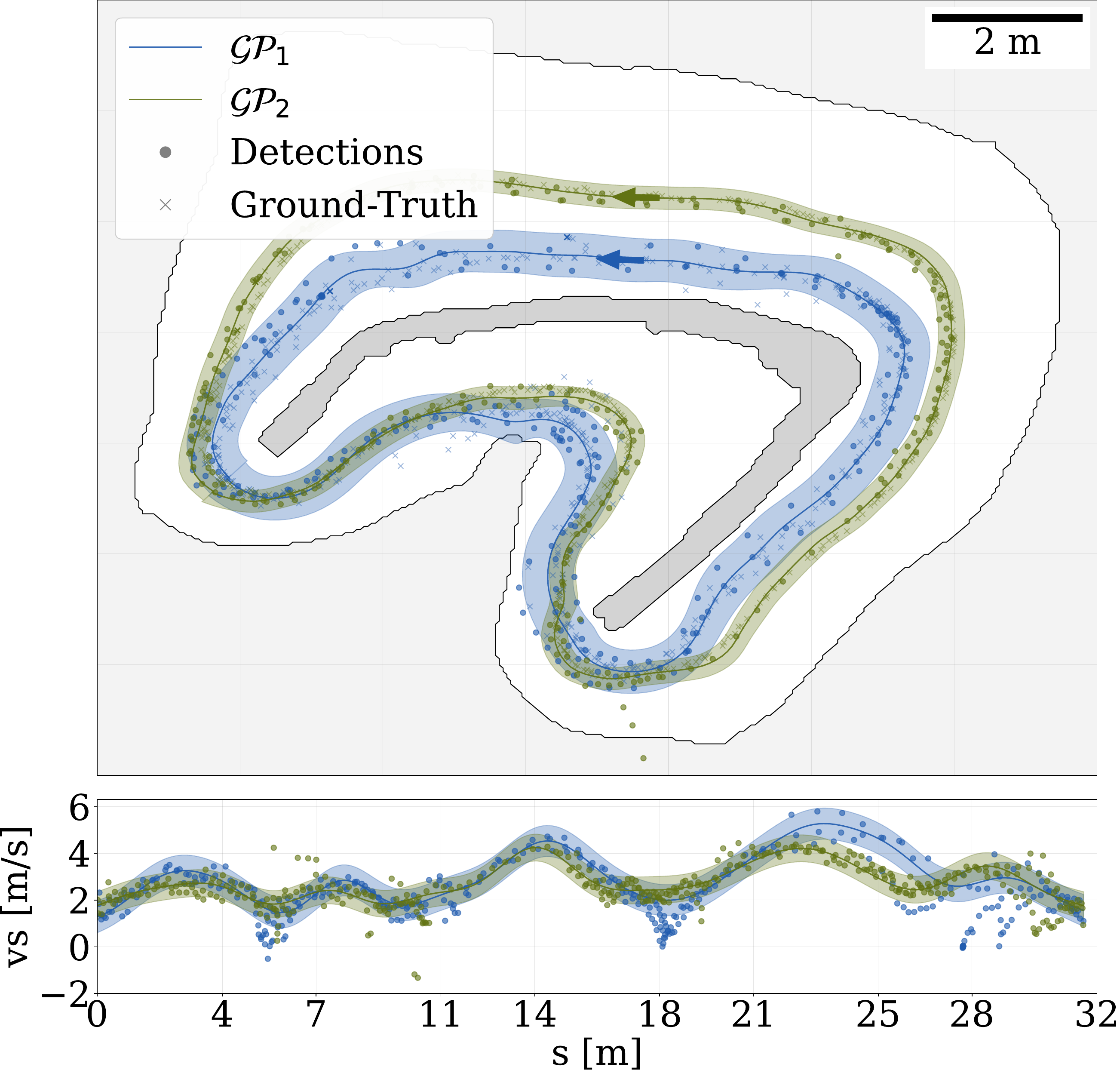}
    \caption{Illustration of the \gls{gpr} for two opponents. $Opp_{1/2}$ are depicted in blue and green respectively. The regressed spatial trajectory $\mathcal{GP}^{d}_{1/2}$ is shown in the top part of the figure and in the bottom the predicted velocity $\mathcal{GP}^{v_s}_{1/2}$ for both opponents. Observations are denoted as circles and ground-truth measurements as $\times$.}
    \label{fig:gp}
\end{figure}

\Cref{fig:gp} demonstrates a qualitative example of how the tracker is capable of keeping track of the \glspl{id} of two opponent vehicles $Opp_1$ in blue and $Opp_2$ in green.

\begin{table*}[!htb]
    \centering
    \begin{adjustbox}{max width=\textwidth}
    \begin{tabular}{l|c|c|c|c|c|c|c|c}
        \toprule
        \textbf{Planner} & \multicolumn{2}{c|}{\textbf{Racing Line}} & \multicolumn{2}{c|}{\textbf{Shortest Path}} & \multicolumn{2}{c|}{\textbf{Centerline}} & \multicolumn{2}{c}{\textbf{Reactive}}\\
        \midrule
        & $\mathcal{R}_{ot/c}[\%] \uparrow$ & $\mathcal{S}_{max}[\%] \uparrow$ & $\mathcal{R}_{ot/c}[\%] \uparrow$ & $\mathcal{S}_{max}[\%] \uparrow$ & $\mathcal{R}_{ot/c}[\%] \uparrow$ & $\mathcal{S}_{max}[\%] \uparrow$ & $\mathcal{R}_{ot/c}[\%] \uparrow$ & $\mathcal{S}_{max}[\%] \uparrow$\\
        \midrule
        \acrshort{pspliner} & \textbf{83.3} & \textbf{77.1} & 71.4 & 77.8 & \textbf{100} & 76.4 & 71.4 & \textbf{75.0}\\
        \acrshort{mpspliner} & \textbf{83.3} & \textbf{77.1} & \textbf{83.3} & \textbf{77.8} & \textbf{100} & \textbf{77.1} & \textbf{100} & \textbf{75.0} \\
        \bottomrule    
    \end{tabular}
    \end{adjustbox}
    \caption{A quantitative analysis of the maximal possible speed for a successful overtake and the success rate. $\mathcal{S}_{max}$ represents the ratio of the \textit{ego} agent's lap time to the maximum lap time of the opponent at which an overtake remains successful. $\mathcal{R}_{ot/c}$ denotes the rate of successful overtakes. The experiments were performed on four different racing behaviors of the opponent as defined in \cref{sec:scenario} until 5 successful overtakes were recorded.}
    \label{tab:single_ot}
\end{table*}

\subsection{Multi-Opponent Motion Prediction}\label{subsec:meth_prediction}
With the multi-opponent tracker described in \Cref{subsec:meth_tracker}, \gls{mpspliner} can simultaneously track multiple opponents. Similar to \gls{pspliner}, the planner follows a structured three-step process for overtaking:

\begin{enumerate}[I] \item \textbf{Trajectory Regression:} Utilize multiple \glspl{gp} to model opponent trajectories.
\item \textbf{Future \acrshort{roc} Computation:} Identify the closest opponent and predict the \gls{roc} ahead.
\item \textbf{Overtaking within the \acrshort{roc}:} Compute a collision-free overtaking trajectory within the \gls{roc}. If additional opponents obstruct the path, their positions are considered spatially.
\end{enumerate}

\subsubsection{Trajectory Regression}
Following the approach in \cite{pspliner}, opponent trajectories are modeled using \glspl{gp} with \gls{gpr}. \Cref{fig:gp} illustrates the trajectory regression for two opponents, $Opp_1$ (blue) and $Opp_2$ (green), where separate \glspl{gp} are employed for spatial ($\mathcal{GP}^d$) and velocity ($\mathcal{GP}^{v_s}$) predictions. The \gls{gp} effectively smooths noisy detections, yielding continuous trajectories.

As in \gls{pspliner}, opponents must be trailed for one lap to construct their respective \glspl{gp}. However, when constructing $\mathcal{GP}_1$ for $Opp_1$, $Opp_2$ may frequently be out of \gls{los}, either due to occlusion by $Opp_1$ or track curvature. With the multi-tracker’s ability to \gls{reid} opponents, $\mathcal{GP}_2$ for $Opp_2$ is built simultaneously with $\mathcal{GP}_1$, ensuring trajectory estimation for multiple opponents even under occlusion.

\subsubsection{Future \acrshort{roc} Computation}
The \gls{roc} is computed based on the closest opponent’s $\mathcal{GP}$ to enable spatiotemporal decision-making, preventing premature overtaking attempts \cite{pspliner}. This is achieved by integrating the opponent's longitudinal position according to the velocity estimated with the $\mathcal{GP}^{v_s}$ of the opponent. If the \textit{ego}-agent has a speed advantage, the integration predicts a spatiotemporal overlap in the racing trajectory, defining the \gls{roc}, following the convention of \cite{pspliner}.

Only a single \gls{roc} is computed, rather than one per opponent, as empirical analysis showed that computing multiple \glspl{roc} was computationally expensive without significant benefit. The likelihood of overlapping \glspl{roc} was minimal, making a single \gls{roc} sufficient for effective planning.

\subsubsection{Overtaking within the \acrshort{roc}}
The overtaking trajectory within the \gls{roc} is computed following the approach in \cite{pspliner}. A sequential quadratic programming optimization problem refines the ego vehicle's trajectory by adjusting its lateral position while ensuring smoothness, respecting track boundaries, and accounting for other opponents spatially if present. Safety is enforced by maintaining a minimum clearance from opponents and limiting curvature to ensure the maneuver remains physically feasible. As a follow up work, integration with a trajectory computation that ensures full feasibility, such as with kinematic-\gls{mpc}, could yield beneficial results \cite{hu2025fsdp}.

\begin{figure*}[!htb]
    \includegraphics[width=\textwidth]{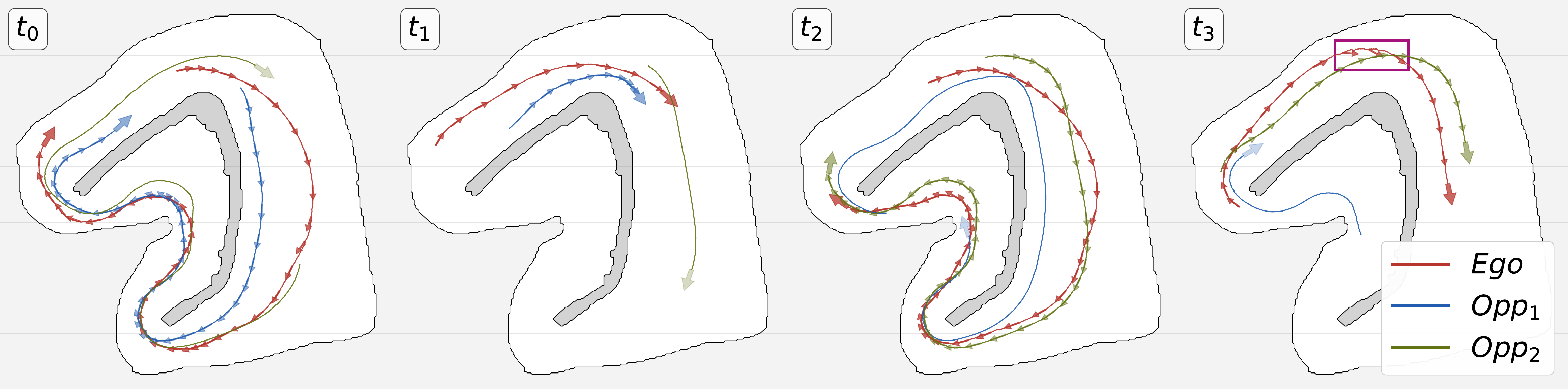}
    \caption{Overview of the different timesteps of the overtaking algorithm. At $t_0$ the \emph{ego} agent is trailing the first opponent to build its $\mathcal{GP}$. At $t_1$ the agent has successfully overtaken the first opponent and starts trailing the second opponent until $t_2$. At $t_3$ it successfully overtakes the second opponent, the box in $t_3$ highlights the high slip-angle of the \emph{ego} agent, as the heading does not align with the trajectory.} 
    \label{fig:qualitative_ots}
\end{figure*}

\section{Experimental Results}
All presented results have been evaluated on a physical 1:10 scaled open-source autonomous racing car \cite{okelly2020f1tenth} and implemented within an open-source autonomy stack to facilitate reproducibility \cite{forzaeth}. The proposed \gls{mpspliner} overtaking algorithm is compared to the single-opponent \gls{pspliner}. To enable a fair comparison, \gls{mpspliner} is first evaluated in a single-opponent scenario to allow direct performance benchmarking against \gls{pspliner}. Subsequently, its capabilities are demonstrated in a multi-opponent scenario, showcasing its effectiveness in handling multiple opponents. The evaluation is structured as follows: the tracking performance is analyzed in \Cref{subsec:tracker_eval}, followed by an assessment of overtaking performance in \Cref{subsec:overtake_eval}.  

\subsection{Opponent Tracking Evaluation}\label{subsec:tracker_eval}
The multi-opponent \gls{reid}-capable tracker described in \Cref{subsec:meth_tracker} is evaluated against the single-opponent \emph{ForzaETH} system \cite{forzaeth}.

To enable a direct comparison, both trackers are tested in a controlled scenario where the ego vehicle and an opponent follow the same racing line, with the opponent moving at a lower speed. The vehicles complete multiple laps while opponent localization data is recorded as ground truth data. The recorded data is then processed offline to evaluate tracking accuracy using both \gls{pspliner} and \gls{mpspliner}.

The top half of \Cref{tab:tracking_metrics} demonstrates the evaluated single-opponent tracking results, where the proposed \gls{mpspliner} tracker achieves a positional error of \SI{0.12}{\metre} over 1264 detections, compared to \SI{0.14}{\metre} over 1245 detections for the single-opponent tracker. This demonstrates that the multi-opponent tracker not only matches the accuracy of the single-opponent tracker in single-opponent scenarios but even slightly outperforms it. The same metrics are presented for the scenario with two opponents in the lower half. Having to keep track of two opponents now, the position \gls{rmse} increases by only \SI{0.06}{\metre}, while the velocity \gls{rmse} is even lower at \SI{0.44}{\metre\per\second} than in the single-opponent case. The \gls{tpr} for two opponents is slightly reduced to 91.2\% compared to one opponent but remains high. This showcases that the tracker can handle multiple oppnents.
\begin{table}[!htb] 
    \centering 
    \resizebox{\columnwidth}{!}{%
    \begin{tabular}{l|c|c|c|c|c}
    \toprule
    \textbf{Tracker} & \textbf{\# Opp} & \bm{$e_{pos}$}$[m]\downarrow$ & \bm{$e_{vel}$}$[m/s]\downarrow$ &  \textbf{\acrshort{tpr}}$[\%]\uparrow$ & \textbf{\acrshort{fdr}}$[\%]\downarrow$ \\
    \midrule
    ForzaETH & 1 & 0.14 & 1.07 & 97.27 & \textbf{0.0} \\
    \gls{mpspliner} & 1 &\textbf{0.12} & \textbf{0.54} & \textbf{98.75} & \textbf{0.0}  \\
    \midrule
    \gls{mpspliner} & 2 & \textbf{0.18} & \textbf{0.44} & \textbf{91.2} & \textbf{0.0}  \\
    \bottomrule
    \end{tabular}%
    }
    \caption{Quantitative comparison of tracking performance between the baseline of the \emph{ForzaETH} racing stack  and our proposed algorihtm \gls{mpspliner}. Metrics include \acrshort{rmse} of the position $e_{pos}$ and velocity $e_{vel}$, \acrshort{tpr} in percentage, and \acrshort{fdr} in percentage. The top half represents tracking metrics against a single opponent, while the bottom half simultaneously handles two opponents.}
    \label{tab:tracking_metrics}
\end{table}



\subsection{Multi-Opponent Overtaking Evaluation}\label{subsec:overtake_eval}
The performance of the proposed multi-opponent capable \gls{mpspliner} is evaluated first by comparing it to the original \gls{pspliner} in a single-opponent scenario. Additionally, \gls{mpspliner} is assessed in a multi-opponent setting to demonstrate its effectiveness in handling multiple opponents simultaneously.

\subsubsection{One-Opponent Scenario}
\label{sec:scenario}
In each experiment, the \emph{ego} agent performs five successful overtakes, while the number of crashes is recorded to compute the overtake success rate, following the methodology in \cite{pspliner}. The \emph{ego}-agent follows the minimum curvature racing line trajectory, while the opponent follows different racing lines at the highest possible speed that still allows for successful overtakes, with the speed scaler \( \mathcal{S}_{max}:= \mathcal{T}_{ego} / \mathcal{T}_{opp} \) computed from lap times on an unobstructed track. Each experiment continued until five successful overtakes (\( \mathcal{N}_{ot} = 5 \)), logging number of crashes (\( \mathcal{N}_{c}\)) to compute the success rate $\mathcal{R}_{ot/c}:=\frac{\mathcal{N}_{ot}}{\mathcal{N}_{ot}+\mathcal{N}_{c}} \in [0,1]$. The \emph{Opp} agent’s behavior was tested under four conditions:  

\begin{enumerate}[I]
    \item \textbf{Racing Line:} Follows the \emph{ego} agent's trajectory, forcing the \emph{ego} agent to deviate from the optimal racing line.
    \item \textbf{Shortest Path:} Takes the shortest path, which overlaps with the \emph{ego} racing line at corner apexes.
    \item \textbf{Centerline:} The \emph{Opp} tracks the centerline which intersects the \emph{ego} agent racing line.
    \item \textbf{Reactive Opponent:} The \emph{Opp} maps \gls{lidar}-based sensor information directly to control actions, by steering into the largest observed gap \cite{ftg}, leading to unpredictable maneuvers and non-deterministic driving behavior.
\end{enumerate}

All opponents, except the reactive opponent type, are non-cooperative, meaning that they do not adapt their trajectories with respect to the ego opponent overtaking. \Cref{tab:single_ot} shows that the proposed \gls{mpspliner} not only matches but outperforms the single-opponent-only \gls{pspliner} in the single-opponent scenario. On average, \gls{mpspliner} improves the overtake success rate \( \mathcal{R}_{ot/c} \) by 10.13 \%-points across all tested \emph{Opp} racing behaviors, increasing from 81.53\% (\gls{pspliner}) to 91.65\% (\gls{mpspliner}), achieving overtakes with fewer crashes. Furthermore, it maintains the same maximum speed scaler \( \mathcal{S}_{max} \), increasing slightly from 76.58\% (\gls{pspliner}) to 76.75\% (\gls{mpspliner}), with an improvement of 0.175 \%-points.

\subsubsection{Multi-Opponent Scenario}  
The time required to overtake 0, 1, and 2 opponents is measured. In the zero-opponent case, only the lap time is recorded, serving as a baseline. When overtaking 1 or 2 opponents, the total time required to complete all overtakes is measured. Each experiment is conducted three times, and the average lap time \( \mu_{\mathcal{T}} \) and the total race time \( \sum \mathcal{T} \) are computed. An Opponent driving on the shortest path at a speed scaler of \( \mathcal{S}_{Opp_1}:= 77.8\% \) is in the front followed by an opponent driving on the centerline with a speed scaler of \( \mathcal{S}_{Opp_2}:= 77.1\% \).

\begin{table}[ht]
    \centering
    \begin{adjustbox}{max width=\columnwidth}
    \begin{tabular}{l|c|c|c|c}
        \toprule
        \textbf{OT Planner} & \textbf{\# Opponents} & \textbf{\# Overtakes} & \bm{$\mu_{\mathcal{T}} [s] \downarrow$} & \bm{$\sum \mathcal{T}[s] \downarrow$} \\ 
        \midrule
        \acrshort{mpspliner} & 0 & 0 & 7.2 & 7.2 \\ \midrule
        \acrshort{mpspliner} & 1 & 1 & 9.25 & 18.5\\ \midrule
        \acrshort{mpspliner} & 2 & 2 & 9.275 & 37.1 \\
        \bottomrule
    \end{tabular}
    \end{adjustbox}
    \caption{Analysis of the \emph{ego} agent's lap time with an increased number of overtakes. $\mu_\mathcal{T}$ represents the mean lap time in seconds, while $\sum \mathcal{T}$ indicates the total racetime in seconds.}
    \label{tab:multi_ot}
\end{table}

\Cref{tab:multi_ot} reports the quantitative results, showing that when overtaking two opponents, the average lap time remains nearly identical to the single-opponent case (\(\mu^{Opp1}_{\mathcal{T}} = \SI{9.25}{\second}\), \(\mu^{Opp2}_{\mathcal{T}} = \SI{9.275}{\second}\)), indicating that \gls{mpspliner} efficiently handles multiple opponents without significant overhead. Furthermore, the cumulative time to perform two overtakes, \(\sum \mathcal{T}_{Opp2} = \SI{37.1}{\second}\), is nearly equal to twice the time required for a single-overtake scenario, \(2\sum \mathcal{T}_{Opp1} = 2 \cdot \SI{18.5}{\second} = \SI{37.0}{\second}\). This highlights that no additional time is wasted, demonstrating that \gls{mpspliner} effectively learns \(\mathcal{GP}_2\) while overtaking \( Opp_1 \), leading to highly efficient multi-opponent overtaking.  

\Cref{fig:qualitative_ots} visually supports the quantitative results of \Cref{tab:multi_ot}, illustrating the \textit{ego} vehicle's overtaking trajectories against two opponents over time, progressing from left to right. Initially, the \emph{ego} agent trails behind \( Opp_1 \), which follows the shortest path, while positioning itself for the overtake on the minimum curvature racing line. Meanwhile, \( Opp_2 \) is further ahead, tracking the centerline  ($t_0$). The \emph{ego} agent first overtakes \( Opp_1 \) on the outside of the corner ($t_1$) before catching up to \( Opp_2 \) and trailing closely ($t_2$). Finally, the \emph{ego} agent executes an inside overtake on \( Opp_2 \), leveraging spatiotemporal knowledge of the opponent's trajectory ($t_3$). Note that the high slip-angle of the \emph{ego} agent at the apex of the \( Opp_2 \) overtake (highlighted box in $t_3$ of \Cref{fig:qualitative_ots}) indicates that it is operating at the limits of friction --- due to the heading not aligning with the driven path --- demonstrating the aggressiveness of the maneuver.

\subsection{Computational Results}
\Cref{tab:compute} presents the computational requirements of the proposed \gls{mpspliner} in comparison to other commonly used overtaking planners. From a computational perspective, the \emph{Spliner} method demonstrates the lowest \gls{cpu} utilization at 6.16\% and the lowest latency at \SI{1.04}{\milli\second}. However, since \emph{Spliner} considers only spatial information and performs significantly worse in terms of safety and effective overtaking velocity, as demonstrated in \cite{pspliner}, the \gls{pspliner} method is the preferred baseline. 

Thus, \gls{mpspliner} is compared primarily to \gls{pspliner}. In terms of computational cost, \gls{mpspliner} exhibits a \SI{31.03}{\percent} \gls{cpu} utilization, which is 7.15\%-points higher than \gls{pspliner}. However, given its ability to handle an arbitrary number of opponents, this increase in computational demand is justified. Additionally, \gls{mpspliner} maintains a latency of \SI{16.12}{\milli\second}, which is nearly identical to that of \gls{pspliner}, ensuring real-time feasibility.  

\begin{table}[htb] 
    \centering 
    \resizebox{\columnwidth}{!}{%
    \begin{tabular}{l|cc|cc|cc}
    \toprule
    \textbf{Planner} & \multicolumn{2}{c|}{\textbf{\acrshort{cpu} [\%]$\downarrow$}} & \multicolumn{2}{c|}{\textbf{Mem [MB]}$\downarrow$} & \multicolumn{2}{c}{\textbf{Latency [ms]}$\downarrow$} \\
    & $\mu_{cpu}$ & $\sigma_{cpu}$ & $\mu_{mem}$ & $\sigma_{mem}$ & $\mu_{t}$ & $\sigma_{t}$ \\
    \midrule
    Frenet & 40.26 & 8.62 & \textbf{36.51} & 2.99 & 12.04 & 30.18 \\
    \gls{gbo} & 56.36 & 22.95 & 147.31 & \textbf{0.01} & 18.01 & 12.25 \\
    Spliner & \textbf{6.16} & \textbf{4.26} & 72.18 & 0.45 & \textbf{1.04} & \textbf{2.62 }\\
    \gls{pspliner} & 23.88 & 31.87 & 112.05 & 0.98 & 15.31 & 37.61 \\
    \acrshort{mpspliner} & 31.03 & 30.57 & 112.09 & 0.83 & 16.12 & 47.77 \\
    \bottomrule
    \end{tabular}%
    }
    \caption{Computational performance of the evaluated planners on an \texttt{AMD Ryzen 7 PRO 7840U}, measured using the \textit{psutil} tool to log \gls{cpu} and memory utilization, as well as the computational time for a single planning loop. Note that \gls{cpu} utilization is reported per core, meaning values can exceed 100\% on a multi-core processor. The mean and standard deviation are denoted by $\mu$ and $\sigma$.}  
    \label{tab:compute}
\end{table}

\section{Conclusion}  
This work presents \gls{mpspliner}, a data-driven overtaking planner that enables multi-opponent racing by leveraging spatiotemporal opponent information for effective overtaking. The key enabler of this approach is the extension of the opponent tracking system, which supports an arbitrary number of opponents through a \gls{sort}-inspired, \gls{kf}-based tracker with \gls{reid} capabilities. The proposed multi-opponent tracker not only matches the performance of the single-opponent tracker but also surpasses it in terms of positional \gls{rmse}, achieving \SI{0.12}{\metre} and velocity \gls{rmse} of \SI{0.54}{\meter/\second}. This enhancement allows for the simultaneous \gls{gpr} of multiple opponents, providing the necessary spatiotemporal information for optimized decision-making. As a result, \gls{mpspliner} achieves an overtaking success rate of up to 91.65\%, representing a 10.13\%-point improvement over the previous \gls{sota} \gls{pspliner}.

\bibliographystyle{IEEEtran}
\bibliography{main}

\begin{thebibliography}{10}
\providecommand{\url}[1]{#1}
\csname url@samestyle\endcsname
\providecommand{\newblock}{\relax}
\providecommand{\bibinfo}[2]{#2}
\providecommand{\BIBentrySTDinterwordspacing}{\spaceskip=0pt\relax}
\providecommand{\BIBentryALTinterwordstretchfactor}{4}
\providecommand{\BIBentryALTinterwordspacing}{\spaceskip=\fontdimen2\font plus
\BIBentryALTinterwordstretchfactor\fontdimen3\font minus \fontdimen4\font\relax}
\providecommand{\BIBforeignlanguage}[2]{{%
\expandafter\ifx\csname l@#1\endcsname\relax
\typeout{** WARNING: IEEEtran.bst: No hyphenation pattern has been}%
\typeout{** loaded for the language `#1'. Using the pattern for}%
\typeout{** the default language instead.}%
\else
\language=\csname l@#1\endcsname
\fi
#2}}
\providecommand{\BIBdecl}{\relax}
\BIBdecl

\bibitem{betz_weneed_ar}
A.~Wischnewski, M.~Geisslinger, J.~Betz, T.~Betz, F.~Fent, A.~Heilmeier, L.~Hermansdorfer, T.~Herrmann, S.~Huch, P.~Karle \emph{et~al.}, ``Indy autonomous challenge-autonomous race cars at the handling limits,'' in \emph{12th International Munich Chassis Symposium 2021: chassis. tech plus}.\hskip 1em plus 0.5em minus 0.4em\relax Springer, 2022, pp. 163--182.

\bibitem{ar_survey}
J.~Betz, H.~Zheng, A.~Liniger, U.~Rosolia, P.~Karle, M.~Behl, V.~Krovi, and R.~Mangharam, ``Autonomous vehicles on the edge: A survey on autonomous vehicle racing,'' \emph{IEEE Open Journal of Intelligent Transportation Systems}, vol.~3, pp. 458--488, 2022.

\bibitem{forzaeth}
N.~Baumann, E.~Ghignone, J.~K{\"u}hne, N.~Bastuck, J.~Becker, N.~Imholz, T.~Kr{\"a}nzlin, T.~Y. Lim, M.~L{\"o}tscher, L.~Schwarzenbach \emph{et~al.}, ``Forzaeth race stack—scaled autonomous head-to-head racing on fully commercial off-the-shelf hardware,'' \emph{Journal of Field Robotics}, 2024.

\bibitem{pspliner}
N.~Baumann, E.~Ghignone, C.~Hu, B.~Hildisch, T.~H{\"a}mmerle, A.~Bettoni, A.~Carron, L.~Xie, and M.~Magno, ``Predictive spliner: Data-driven overtaking in autonomous racing using opponent trajectory prediction,'' \emph{IEEE Robotics and Automation Letters}, 2024.

\bibitem{mpcc}
A.~Liniger, A.~Domahidi, and M.~Morari, ``Optimization-based autonomous racing of 1:43 scale rc cars,'' \emph{Optimal Control Applications and Methods}, vol.~36, no.~5, pp. 628--647, 2015.

\bibitem{ftg}
V.~Sezer and M.~Gokasan, ``A novel obstacle avoidance algorithm: “follow the gap method”,'' \emph{Robotics and Autonomous Systems}, vol.~60, no.~9, pp. 1123--1134, 2012.

\bibitem{frenetplanner}
M.~Werling, J.~Ziegler, S.~Kammel, and S.~Thrun, ``Optimal trajectory generation for dynamic street scenarios in a frenet frame,'' in \emph{2010 IEEE International Conference on Robotics and Automation}.\hskip 1em plus 0.5em minus 0.4em\relax IEEE, 2010, pp. 987--993.

\bibitem{gbo}
T.~Stahl, A.~Wischnewski, J.~Betz, and M.~Lienkamp, ``Multilayer graph-based trajectory planning for race vehicles in dynamic scenarios,'' in \emph{2019 IEEE Intelligent Transportation Systems Conference (ITSC)}.\hskip 1em plus 0.5em minus 0.4em\relax IEEE, 2019, pp. 3149--3154.

\bibitem{norfair}
\BIBentryALTinterwordspacing
J.~Alori, A.~Descoins, javier, F.~Lezama, KotaYuhara, D.~Fernández, A.~Castro, fatih, David, R.~C. Linares, F.~Kurucz, B.~Ríos, shafu.eth, K.~Nar, D.~Huh, and Moises, ``tryolabs/norfair: v2.2.0,'' Jan. 2023. [Online]. Available: \url{https://doi.org/10.5281/zenodo.7504727}
\BIBentrySTDinterwordspacing

\bibitem{okelly2020f1tenth}
M.~O’Kelly, H.~Zheng, D.~Karthik, and R.~Mangharam, ``F1tenth: An open-source evaluation environment for continuous control and reinforcement learning,'' in \emph{NeurIPS 2019 Competition and Demonstration Track}.\hskip 1em plus 0.5em minus 0.4em\relax PMLR, 2020, pp. 77--89.

\bibitem{boosting}
H.~Grabner, M.~Grabner, and H.~Bischof, ``Real-time tracking via on-line boosting.'' in \emph{Bmvc}, vol.~1, no.~5.\hskip 1em plus 0.5em minus 0.4em\relax Citeseer, 2006, p.~6.

\bibitem{kcf}
J.~F. Henriques, R.~Caseiro, P.~Martins, and J.~Batista, ``High-speed tracking with kernelized correlation filters,'' \emph{IEEE transactions on pattern analysis and machine intelligence}, vol.~37, no.~3, pp. 583--596, 2014.

\bibitem{medianflow}
Z.~Kalal, K.~Mikolajczyk, and J.~Matas, ``Forward-backward error: Automatic detection of tracking failures,'' in \emph{2010 20th international conference on pattern recognition}.\hskip 1em plus 0.5em minus 0.4em\relax IEEE, 2010, pp. 2756--2759.

\bibitem{xiao2024motiontrack}
C.~Xiao, Q.~Cao, Y.~Zhong, L.~Lan, X.~Zhang, Z.~Luo, and D.~Tao, ``Motiontrack: Learning motion predictor for multiple object tracking,'' \emph{Neural Networks}, vol. 179, p. 106539, 2024.

\bibitem{sort}
A.~Bewley, Z.~Ge, L.~Ott, F.~Ramos, and B.~Upcroft, ``Simple online and realtime tracking,'' in \emph{2016 IEEE international conference on image processing (ICIP)}.\hskip 1em plus 0.5em minus 0.4em\relax IEEE, 2016, pp. 3464--3468.

\bibitem{davide}
D.~Plozza, S.~Marty, C.~Scherrer, S.~Schwartz, S.~Zihlmann, and M.~Magno, ``Autonomous navigation in dynamic human environments with an embedded 2d lidar-based person tracker,'' in \emph{2024 IEEE Sensors Applications Symposium (SAS)}, 2024, pp. 1--6.

\bibitem{mpcot}
E.~L. Zhu, F.~L. Busch, J.~Johnson, and F.~Borrelli, ``A gaussian process model for opponent prediction in autonomous racing,'' in \emph{2023 IEEE/RSJ International Conference on Intelligent Robots and Systems (IROS)}, 2023, pp. 8186--8191.

\bibitem{story_of_modelmismatch}
\BIBentryALTinterwordspacing
A.~Liniger, ``Pushing the limits of friction: A story of model mismatch,'' 2021, iCRA21 Autonomous Racing. [Online]. Available: \url{https://www.youtube.com/watch?v=_rTawyZghEg&t=136s}
\BIBentrySTDinterwordspacing

\bibitem{amin2022}
\BIBentryALTinterwordspacing
D.~E. Amin, K.~Priandana, and M.~K.~D. Hardhienata, ``Development of adaptive line tracking breakpoint detection algorithm for room sensing using lidar sensor,'' \emph{International Journal of Advanced Computer Science and Applications}, vol.~13, no.~7, 2022. [Online]. Available: \url{http://dx.doi.org/10.14569/IJACSA.2022.0130732}
\BIBentrySTDinterwordspacing

\bibitem{hu2025fsdp}
C.~Hu, J.~Huang, W.~Mao, Y.~Fu, X.~Chi, H.~Qin, N.~Baumann, Z.~Liu, M.~Magno, and L.~Xie, ``Fsdp: Fast and safe data-driven overtaking trajectory planning for head-to-head autonomous racing competitions,'' \emph{arXiv preprint arXiv:2503.06075}, 2025.

\end{thebibliography}

\end{document}